\documentclass[10pt,twocolumn,letterpaper]{article}

\PassOptionsToPackage{table}{xcolor}

\usepackage[pagenumbers]{wacv} 

\usepackage{amsmath,amsfonts,bm}

\def\eqref#1{equation~\ref{#1}}

\def\1{\bm{1}}

\def\va{{\bm{a}}}

\def\vc{{\bm{c}}}

\def\vh{{\bm{h}}}

\def\vp{{\bm{p}}}
\def\vq{{\bm{q}}}

\def\vx{{\bm{x}}}

\DeclareMathAlphabet{\mathsfit}{\encodingdefault}{\sfdefault}{m}{sl}
\SetMathAlphabet{\mathsfit}{bold}{\encodingdefault}{\sfdefault}{bx}{n}

\DeclareMathOperator*{\argmin}{arg\,min}

\usepackage{subcaption}

\definecolor{wacvblue}{rgb}{0.21,0.49,0.74}
\usepackage[pagebackref,breaklinks,colorlinks,allcolors=wacvblue]{hyperref}

\usepackage[utf8]{inputenc} 
\usepackage[T1]{fontenc}    
\usepackage{url}            
\usepackage{booktabs}       
\usepackage{amsfonts}       
\usepackage{afterpage}
\usepackage{nicefrac}       
\usepackage{microtype}      
\usepackage{comment}        
\usepackage{amsmath}
\usepackage{multirow}
\usepackage{adjustbox}
\usepackage{bm}
\usepackage{amsthm}
\usepackage{hhline}
\usepackage{amssymb}
\usepackage{makecell}
\usepackage{mathtools}
\usepackage{color}
\usepackage{MnSymbol}
\usepackage{graphicx}
\usepackage{arydshln}
\usepackage{framed}
\usepackage[font=small]{caption}
\usepackage[scientific-notation=true]{siunitx}
\usepackage{wrapfig}
\usepackage{lipsum}
\usepackage{sidecap}
\usepackage{pifont}
\usepackage[flushleft]{threeparttable}
\usepackage{diagbox}
\usepackage{enumitem}
\usepackage{tabularx}
\usepackage[parfill]{parskip}
\usepackage{bbm}
\usepackage{cases}
\usepackage{subcaption}
\usepackage{algorithm}
\usepackage{longtable}
\usepackage{algorithmic}
\usepackage{xspace}
\usepackage{tikz}
\usepackage{newtxtext}
\usepackage{colortbl}
\usepackage{listings}
\usepackage{tcolorbox}

\newcommand{\model}{\textsc{VAP}\xspace}

\def\wacvPaperID{1100} 
\def\confName{WACV}
\def\confYear{2026}

\title{Video Active Perception: Effective Inference-Time \\ Long-Form Video Understanding with Vision-Language Models}

\author{Martin Q. Ma$^{1}$
\and 
Willis Guo$^{1}$\thanks{Equal contributions.}
\and
Aditya Agrawal$^{1*}$
\and
Ankit Gupta$^{1*}$
\and
Paul Pu Liang$^{2}$
\and
Ruslan Salakhutdinov$^{1}$
\and
Louis-Philippe Morency$^{1}$
\and
\\
$^1$Carnegie Mellon University \hspace{5em} 
$^2$MIT
}

\begin{document}
\maketitle
\begin{abstract}
Large vision-language models (VLMs) have advanced multimodal tasks such as video question answering (QA). However, VLMs face the challenge of selecting frames effectively and efficiently, as standard uniform sampling is expensive and performance may plateau. Inspired by active perception theory, which posits that models gain information by acquiring data that differs from their expectations, we introduce Video Active Perception (\model{}), a training-free method to enhance long-form video QA using VLMs. Our approach treats keyframe selection as data acquisition in active perception and leverages a lightweight text-conditioned video generation model to represent prior world knowledge. Empirically, \model{} achieves state-of-the-art zero-shot results on long-form or reasoning video QA datasets such as EgoSchema, NExT-QA, ActivityNet-QA, IntentQA, and CLEVRER, achieving an increase of up to $5.6 \times$ frame efficiency by frames per question over standard GPT-4o, Gemini 1.5 Pro, and LLaVA-OV. Moreover, \model{} shows stronger reasoning abilities than previous methods and effectively selects keyframes relevant to questions. These findings highlight the potential of leveraging active perception to improve the frame effectiveness and efficiency of long-form video QA.
\end{abstract}    
\section{Introduction}
Multimodal foundational models, particularly large vision-language models (VLMs) \citep{achiam2023gpt, reid2024gemini}, have achieved remarkable results in tasks such as image captioning, text-to-image generation, and video question answering~\citep{liang2024foundations}. Long-form video question answering \citep{xiao2021next, mangalam2024egoschema} stands out as a challenging and intriguing problem. It requires models to reason over complex dynamics, intricate scenes, and subtle visual details across extended time frames. Developing effective solutions for this task is practically important and poses compelling scientific challenges.

However, long videos produce two challenges during VLM inference. The first is \textit{frame effectiveness}: standard VLMs use one-frame-per-second (fps) sampling of frames, which is suboptimal as some video regions can be irrelevant while others can be information-dense and task-relevant \citep{wang2024videotree}. In Table \ref{tab:increasing_num_frames_plateaus}, we show the performance of uniform sampling plateaus on the NeXT-QA dataset, suggesting that the uniformly sampled frames may not be the most effective. The second challenge is \textit{frame efficiency}: processing a single hour of video can produce nearly four million tokens. The prohibitive token amounts hinders the practical deployment of VLMs, where real-world video applications such as autonomous driving, content summarization, and patient monitoring with thousands of hours of video data.

\begin{figure*}
\centering
\includegraphics[scale=0.174]{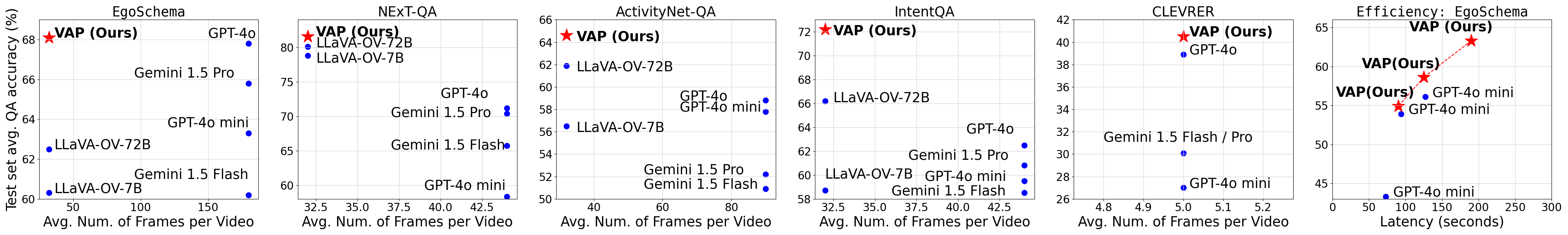}
\caption{\model{} outperforms standard uniform sampling GPT-4o, Gemini 1.5 and LLaVA-OV on EgoSchema, NExT-QA, ActivityNet-QA, IntentQA, and CLEVRER with up to $5.6 \times$ more efficient on frame per question. \model{} also achieves better latency under the same performance with GPT-4o mini compared to standard sampling due to using fewer frames.} 
\label{fig:performance}
\end{figure*}

In this paper, we introduce \textbf{Video Active Perception (\model{})}, a training-free method that improves both the frame effectiveness and efficiency for long-form video question answering, by drawing inspiration from ``active perception''~\citep{bajcsy1988active, aloimonos2013active, bajcsy2018revisiting, tenenbaum1971accommodation, mcarthur1983toward, dijksterhuis2001perception, satchell2021ecological, wu2024network}, which posits that intelligent agents should guide \textit{data acquisition} through \textit{a priori knowledge} of the world. 
Concretely, the data acquisition corresponds to selecting key video frames in videos for VLM inference, while the a priori knowledge is embodied by a lightweight, pre-trained text-conditional video generation model that encodes complex prior visual knowledge. Initial uniform sampled frames, along with questions and possible answers, are the conditional signals to the diffusion model. The latents from the diffusion model are then compared against a separate set of latents encoded from all real frames. The real frames whose latent are the most distinct from the diffusion latents are selected for VLM inference. Unlike previous frame selection methods \citep{wang2024videoagent, fan2024videoagent, wang2024videotree}, \model{} \textit{does not require a captioning model} and operates in a \textit{single selection round} rather than through multi-round selection or complex data structures, providing a simplified, unified approach.

\model{} demonstrates substantial empirical improvements in both performance and frame efficiency across long-form video QA datasets such as EgoSchema \citep{mangalam2024egoschema}, NExT-QA \citep{xiao2021next}, ActivityNet-QA \citep{yu2019activitynet}, IntentQA \citep{li2023intentqa}, and the reasoning dataset CLEVRER \citep{yi2019clevrer}. Performance-wise, by selectively focusing on frames that diverge from prior knowledge, \model{} obtains state-of-the-art zero-shot results with $68.1\%$ on EgoSchema, $81.4\%$ on NExT-QA, $64.6\%$ on ActivityNet-QA, $72.2\%$ on IntentQA, and $40.5\%$ on CLEVRER (Table \ref{tab:main_results}). It also outperforms standard uniform sampling baselines with GPT-4o, Gemini 1.5 Pro, and LLaVA-OV-72B \citep{li2024llava} (Figure \ref{fig:performance}).  For frame efficiency, \model{} achieves up to a $5.6 \times$ increase in efficiency by frames per question with similar or better performance (Figure \ref{fig:performance}). Furthermore, compared to the standard GPT-4o mini, \model{} achieves better efficiency through a lower latency (Figure \ref{fig:performance} and Table \ref{tab:efficiency_same_performance}) with the same performance due to the use of fewer frames. \model{} is both VLM-agnostic and task-agnostic, which makes it applicable in different vision-language models and tasks. Our quantitative analysis reveals that \model{} exhibits stronger visual reasoning capabilities than previous caption-based models on challenging temporal and causal reasoning tasks (Section \ref{sec:better_reasoning_over_prior_art}). Qualitative results illustrate that \model{} effectively selects unseen pivotal frames relevant to the questions for explanatory and counterfactual reasoning (Section \ref{sec:qualitative}). These findings demonstrate the effectiveness of leveraging prior world knowledge from a generation model to enhance both the frame effectiveness and efficiency of long-form video question answering, highlighting the potential of a more intelligent data acquisition strategy for VLM inference.

\section{Related work}
\paragraph{Active perception, active feature and input acquisition.} Active perception \citep{tenenbaum1971accommodation, bajcsy1988active, aloimonos2013active, pulvermuller2010active, bajcsy2018revisiting, satsangi2020maximizing, zaky2020active, zhang2021safe} refers to the theory in which an agent actively acquires its sensory input to optimize perception based on feedback from an a priori knowledge model. Similar ideas include active sensing \citep{ji2007cost, yu2009active, yang2016active, yin2020reinforcement}, active feature acquisition \citep{saar2009active, shim2018joint, lewis2021accurate}, and active data acquisition \citep{kossen2022active, lai2023active}. In this work, we leverage the active perception framework by using a video generation model to represent a priori knowledge and selecting keyframes as a data acquisition process for VLM inference. 

\paragraph{Frame selection methods for inference-time video question-answering.} Long-form video question answering presents significant challenges for large vision-language models (VLMs) due to the computational cost. Several approaches rely on LLMs iteratively reasoning over captions of densely sampled video frames \citep{liao2024videoinsta, ma2024drvideo, goulas2024vidctx, zhang2024llovi, wang2024videotree}. Recently, agent-based methods \citep{yang2024vca, shang2024traveler, fan2024videoagent, wang2024videoagent} have also been proposed, though many of these agents still rely on video frame captions. In addition to video frame captions, other methods leverage external tools like object detection to build a memory module for RAG \citep{luo2024videorag, jeoung2024avua, min2024morevqa} Elsewhere, other methods represent videos as graphs \citep{huang2025mindpalace}, image grids \citep{kim2024image}, or perform spatiotemporally pooled visual features \citep{xu2024slowfastllava}. Among all approaches, many methods rely on video frame captions. However, caption-based methods are sensitive to captioner quality \cite{zhang2024llovi}. Furthermore, captions inevitably lose critical information \cite{ma2024drvideo}, requiring video agents whose iterative workflows incur high latency as our results show. In contrast, our approach requires only a lightweight pre-trained video generation model, eliminating the need for captioning models or complex data structures. By selecting frames in a single round rather than multiple iterations, our method is more simple and efficient. Using the principles of active perception, we focus on frames that diverge most from expectations from the video generation model. 

\section{Video Active Perception}
\begin{figure*}
    \centering
    \includegraphics[width=0.9\textwidth]{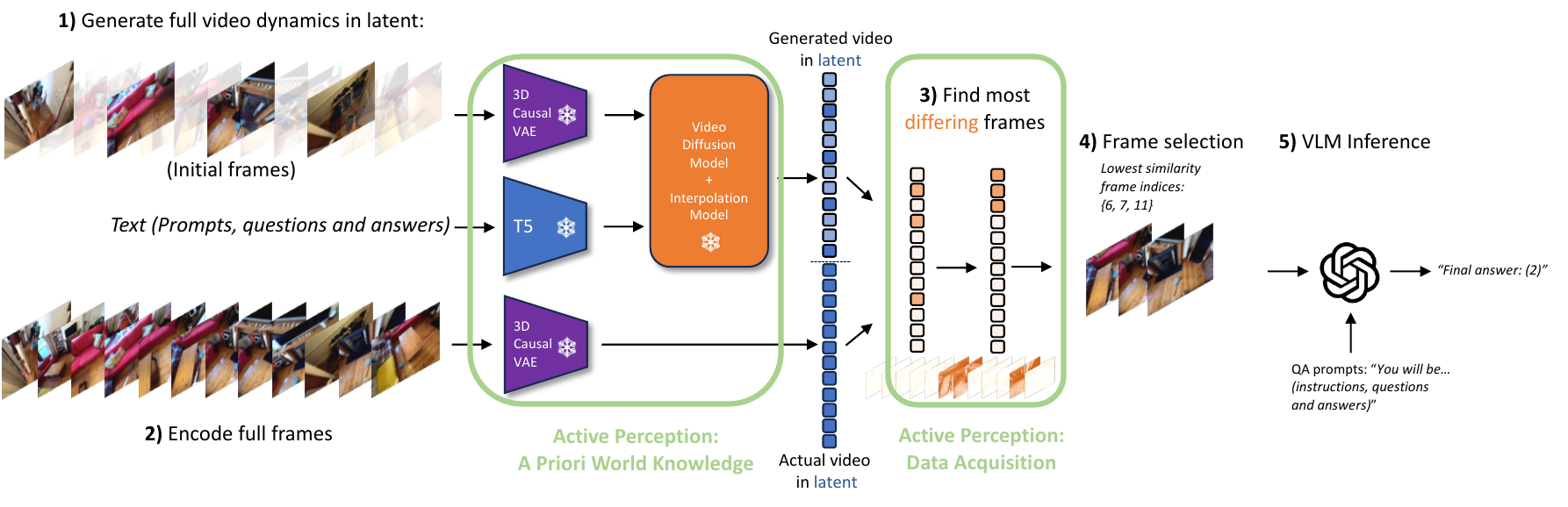}
    \caption{Overview of Video Active Perception model (\model{}). There are two core modules in \model{}: an a priori knowledge model, and a data acquisition process. The a priori knowledge model, which is a conditional diffusion model, generates full video dynamics from a few initial frames and QA information. The data acquisition process compares the generated vs. real video dynamics and finds actual frames that are the most informative based on differences from the expected video dynamics. The selected frames are used for the VLM inference.}
    \label{fig:overview}
\end{figure*}

In this section, we introduce the key technical components of Video Active Perception: a) a priori world knowledge: producing the full video dynamics in the latent space via the lightweight text-conditioned generative model, in Section \ref{sec:generate_unseen}); b) data acquisition: frame selection based on the comparison between the generated and real frames, in Section \ref{sec:frame_selection}; and c) VLM inference: running inference on flagship VLMs from the selected frames, in Sec \ref{sec:inference_from_frames}. 

We formulate training-free, inference-time usage instead of fine-tuning or pre-training large Vision-Language Models (VLMs) for video question answering. We have a long-form video $\vx_{1:T}=(\vx_1, \dots, \vx_T)$ with frames $\vx_i$, and a large total number of frames $T$. The video comes with the question $\vq$, possible answers $a$, and the user-defined prompt to facilitate VLMs. We assume a lightweight pre-trained text-conditioned video generation model $f(\cdot)$, which takes a sequence of $t$ video frames $\vx_{1:t}$, where $t$ is the number of frames and the text prompt $p$, and output latents with full frame numbers $\tilde{\vh}_{1:T}$ (details explained below). Our goal is to select a subset of $K$ frames from $\vx_{1:T}$ to perform inference efficiently with a large VLM $g(\cdot)$. The overview algorithm of our method is given in Algorithm \ref{alg:ap}, and we will break it down into the following sections. 

\subsection{A Priori Knowledge for Generating Full Video Dynamics} \label{sec:generate_unseen}
We present the a priori knowledge module in \model{}: generating video dynamics by producing unseen frames from only a few frames along with the question and answers. \model{} uses a pre-trained, lightweight video generation model to encode the seen frames and texts, and a frame interpolation module to produce the latents corresponding to the unseen frames. Because of the conditioning, the generation model is strictly speaking a conditional predictive prior instead of purely a priori. Then, all latents (seen and generated frames, and text) will be passed to a stack of transformer blocks for better alignment of visual and textual feature spaces. The model output are unpatchified to the original latent shape, and will be fed into a denoiser of a video diffusion model as conditioning signals to sample latent frames. For the generation model, we use CogVideo \citep{hong2022cogvideo} and its updated variation, CogVideoX-5B \citep{yang2024cogvideox}, as the video generation model. CogVideo is an open-source large-scale diffusion transformer for general text-to-video generation. CogVideoX \citep{yang2024cogvideox}, based on CogVideo, and is a state-of-the-art large-scale diffusion transformer model for text-conditioned video generation. We use CogVideoX for encoding due to better compression from pixels to latent spaces, and the recursive interpolation module from CogVideo for frame interpolation. 
We list the key steps below, and the training details of CogVideoX can be found in \cite{yang2024cogvideox} and its codebase.

\paragraph{Uniform sampling initial frames.} First, we uniformly sample a small subset of initial frames from the video as bare bones for generation. These frames are supposed to provide the basic dynamics and visual context of the videos. We sample up to $n=32$ frames in our experiments, a small amount compared to long videos. 

\paragraph{Encoding sampled frames.} We adopt the pre-trained 3D VAE \citep{yu2023language, yang2024cogvideox}, which incorporate 3D convolutions to compress video spatially and temporally to achieve higher compression ratio for improved video reconstruction quality and continuity. The encoder of the 3D VAE each contains four 2 $\times$ downsampling, with both the spatial and temporal dimension in the first rounds and spatial dimension in the last round, achieving a $4 \times 8 \times 8$ compression from pixels to latents. This step compresses enough visual information to generate rich, contextualized unseen frames in latent.

\paragraph{Alignment betweeen vision and text.} The latents of the sampled frames and the interpolated frames are combined by their interpolation ordering, patchified along the spatial dimension. Following \citet{yang2024cogvideox}, a 3D Rotary Position Embedding \citep{su2024roformer}, a relative position encoding that is better than the sinusoidal absolute position encoding \citep{yang2024cogvideox}, is applied to the spatio and temporal dimensions. The text input is encoded using T5 \citep{raffel2020exploring}. The latents of both modality are fed into a stack of diffusion transformer blocks. Following DiT \citep{peebles2023scalable}, we also use the timestep $t$ of the diffusion process as the input to the transformer. Modality-specific adaptive layer norms are applied to both the video and the text latent, which is shown to promote the alignment of feature spaces across modalities \citep{yang2024cogvideox}. A 3D text-video hybrid attention mechanism is used, as introduced in \citet{yang2024cogvideox}. 

\paragraph{Generating latent frames.} Next, we seek to sample (generate) video frames from the video diffusion module \citep{ho2022video} of CogVideoX, but only in latent space. First, the outputs of the last step are unpatchified to restore the original latent shape. Then, we simply sample from the video diffusion model by feeding the aligned visual-textual latents as the conditioning signal. We then leverage the pre-trained frame interpolation model from CogVideo \citep{hong2022cogvideo}, which is based on Real-Time Intermediate Flow Estimation (RIFE) \citep{huang2022real}. The frames in latent space are chunked into frame blocks, and for each block, a frame in latent space is interpolated with guidance of structural similarity index measure (SSIM). SSIM threshold can be adjusted to reach the desired number of frames. For very long videos where the number of frames is large, we leverage a memory bank to cache the latents. The RIFE model is fast and the interpolation is light-weighted, and we regard it as an integral part of the generation model. 

\begin{algorithm}[t]
    \caption{Algorithm of \model{}. }
    \label{alg:ap}
    \begin{algorithmic}[1]
        \REQUIRE Inference video $\vx_{1:T}$, prompt $\vp$, question $\vq$ and answers $\va$, video generation model $f(\cdot)$, and a large Vision-Language Model (VLM) $g(\cdot)$, initial frame number $k$, final frame number $n$.
        \STATE Uniformly sample $k$ initial frames from $\vx_{1:T}$: $\{\vx_i\}_{i \in \mathcal{S}}$, where $\mathcal{S} \subseteq \{1,2,\dots,T\}, |\mathcal{S}|=k\}$
        \STATE Get a set of latents by encoding initial frames and then sampling from the generation model: $\tilde{\vh}_{1:T} \gets f\left(\{\vx_i\}_{i \in \mathcal{S}} | \vq, \va \right)$
        \STATE Get a set of latents from encoding all real frames: $\vh_{1:T} \gets f(\vx_{1:T} | \vq, \va)$
        \STATE Compute the cosine similarity $\vc$: $c_i \gets \vh_i  \cdot \tilde{\vh}_i$
        \STATE Sort and find $n$ indices with lowest similarities: $\mathcal{I}_n \gets \underset{i_1, \dots, i_n}{\argmin}$ $(c_{i})$. 
        \STATE Select $k$ frames by $\mathcal{I}_n$: $\{\vx_i | i \in \mathcal{I}_n\}$
        \STATE Return response from VLM with selected frames: $g\left(\vp, \vq, \{\vx_i | i \in \mathcal{I}_n\}\right)$.
    \end{algorithmic}
\end{algorithm}

\subsection{Data Acquisition by Video Frame Selection}
\label{sec:frame_selection}
The next step is data acquisition, which involves selecting the most informative and ``surprising'' frames. From the active perception principles, these are the frames that diverge the most from the expected (generated) video dynamics produced by the generation model.

\paragraph{Encoding all real frames.} We use the 3D VAE encoder of the same video generation model, CogVideoX, to encode all real frames. Doing so allows us to compare real frames against generated frames in an efficient way, since all encoded frames are in latent space only. The 3D VAE achieves $4 \times 8 \times 8$ compression, and we use a memory bank for long videos from EgoSchema to store the latents. Note that we currently did not further leverage the sampling capabilities from the generation model (i.e., the video diffusion model) to further improve efficiency.

\paragraph{Similarity between real and generated frames.} Given the real frames in the latent space and the generated frames in the latent space, we now acquire the most informative real frames by comparing the two sets of latent frames against each other. In this paper, we use cosine similarity to select the most dissimilar real frames from their generated counterparts. All real frames will have one corresponding generated frame in latent space; therefore, we compute the cosine similarity for each pair. We then sort the cosine similarity list and select the indices of $n=32$ real frames with the lowest similarities to their generated counterparts for VLM inference. In Section \ref{tab:num_frames_ablation} we explore different frame numbers and find $n=32$ sufficiently optimal for our tasks.

\subsection{VLM inference from selected frames} \label{sec:inference_from_frames}
After locating the frame indices, we combine the frames into a set. With this set and an instructional prompt, the question and all possible answers, we use them to run VLM inference. We include the exact formats of the prompts in the Appendix (Section A.1). All VLMs we use (GPT-4o, GPT-4o mini, GPT 1.5 Pro, GPT 1.5 Flash, LLaVA OneVision-7B and LLaVA-OneVision 72B) can take interleaved prompts of images and texts as input. After inference with VLM on each video QA dataset, we gather the generated responses, parse the answers, and evaluate the results.

\section{Experiments}
In this section, we introduce the datasets, evaluation metrics, baselines, implementation details, main results of \model{}, and quantitative and qualitative analyzes.

\subsection{Datasets and Metrics}
\paragraph{EgoSchema.} Egoschema \citep{mangalam2024egoschema} is a dataset with $5,000$ videos, along with an associated question and answer pair. Each question is a multiple-choice question that has a series of five answers that are associated with it. Each video clip is around three minutes long. The videos in EgoSchema cover many different types of human behavior. In order to evaluate the model's performance on EgoSchema, we measure the percentage of predicted multiple choice answer options that match the correct multiple choice answer option. The accuracy of multiple-choice answers are measured by submitting to the EgoSchema leaderboard \citep{egoschema_leaderboard}. 

\paragraph{NExT-QA.} NExT-QA \citep{xiao2021next} is a dataset to study causal and temporal action reasoning in video. It contains $5,440$ videos and about $52,000$ manually annotated question-answer pairs grouped into causal, temporal, and descriptive questions. There are multi-choice QA provides five candidate answers, as well as open-ended QA. This dataset challenges models to truly understand the causal and temporal relationships of the actions throughout the videos. 
\paragraph{ActivityNet-QA.} ActivityNet-QA \citep{yu2019activitynet} consists of 58,000 QA pairs on 5,800 complex web videos derived from the popular ActivityNet \citep{caba2015activitynet} dataset, which contains diverse web videos with two hundred action classes. The average video length is three minutes. We follow \citet{maaz2023video} to use GPT-3.5 to evaluate the open-ended questions.
\paragraph{IntentQA.} IntentQA \citep{li2023intentqa} is a dataset with daily social activities. It contains three types of contexts, situational, contrastive, and commonsense contexts to provide context for intent understanding from videos.
\paragraph{CLEVRER.} CLEVRER \citep{yi2019clevrer} is a collision event-based video dataset that studies the temporal and causal structures behind the videos of simple objects. It includes 20,000 synthetic videos of colliding objects and more than $300,000$ questions and answers. It has four types of questions: descriptive (for example, ``what color''), explanatory (``what’s responsible for''), predictive (``what will happen next''), and counterfactual (``what if''). We report per question the accuracy and submit to the official evaluation server \citep{EvalAI} to get results from the test set.

\begin{table*}[t]
\small
\caption{Comparison of test set accuracy on zero-shot video QA tasks on EgoSchema, NExT-QA, ActivityNet-QA, IntentQA, and CLEVERER datasets (standard baselines results are reproduced). \model{} achieves state-of-the-art results on all five datasets.}  

\centering
\resizebox{\textwidth}{!}{
\begin{tabular}{lrcccccccccccccccccc}
\toprule
\multirow{2}{*}{\textbf{Model}} & \multicolumn{1}{c}{\multirow{2}{*}{\textbf{VLM}}} & \multicolumn{2}{c}{\textbf{EgoSchema}}                                     &  & \multicolumn{4}{c}{\textbf{NExT-QA}}                                                                                &  & {\textbf{ANet-QA}}  & & {{\textbf{IntentQA}}} & & \multicolumn{4}{c}{\textbf{CLEVRER}}                \\ \cmidrule(lr){3-4} \cmidrule(lr){6-9} \cmidrule(lr){10-11} \cmidrule(lr){12-13} \cmidrule(lr){14-19} 
                                & \multicolumn{1}{c}{}                             & \multicolumn{1}{c}{\textbf{Sub.}} & \multicolumn{1}{c}{\textbf{Full}} &  & \multicolumn{1}{c}{\textbf{Tem.}} & \multicolumn{1}{c}{\textbf{Cau.}} & \multicolumn{1}{c}{\textbf{Des.}} & \textbf{Avg.} & & \textbf{Test} & & \multicolumn{1}{c}{{\textbf{Avg.}}} & & \textbf{Des.} & \textbf{Exp.} & \multicolumn{1}{c}{\textbf{Pre.}} & \multicolumn{1}{c}{\textbf{Cou.}} & \multicolumn{1}{c}{\textbf{Avg.}} \\
\midrule
\rowcolor{gray!20} \multicolumn{18}{c}{{\textit{Standard baselines, 1fps uniform sampling}}}   \\

LLaVA-OneVision (repro.) & LLaVA-OV-7B& 62.4 &60.3 & & 76.0 & 79.5&85.5& 79.4 & &56.5 & & 58.7& & - & - & - & - & -\\
LLaVA-OneVision (repro.) & LLaVA-OV-72B& 64.2 & 62.5 & &  77.8&81.2&86.1& 80.9 &  & 61.9 & &{66.2} & &- & - & - & - & - \\
Gemini 1.5 Flash (repro.) &Gemini 1.5 Flash& 63.3 & 60.2 & & 62.9 & 65.9&70.4 & 65.7& &50.9& & 59.5 & & 42.9 & 9.7 & 53.2 & 14.4 & 30.1 \\
Gemini 1.5 Pro (repro.) &Gemini 1.5 Pro& 67.5 & 65.8 & & 66.7 &71.8 &73.2 &70.4 & &52.2& & 60.8 & & 47.9 & 16.3 & 45.1 & 10.6 & 30.0  \\
GPT-4o mini (repro.) &GPT-4o mini& 65.1 &63.3 & &  55.3&57.5&66.7&58.3 &  &57.8 &  &58.5  & & 34.6 &  14.9 & 45.6 & 12.7 & 27.0  \\
GPT-4o (repro.) &GPT-4o& 69.2 & 67.8 & &  67.6&71.9&75.8&71.2 & &58.8 & & 62.5 & & 38.2 & 28.3 & \textbf{65.0} & 24.1 & 38.9 \\

\rowcolor{gray!20} \multicolumn{18}{c}{{\textit{Frame selection baselines}}}   \\

VideoAgent~\citep{wang2024videoagent} &GPT-4 &  60.2&54.1 & &  64.5 & 72.7 & 81.1 & 71.3 & & - & & - & &- &- &- &- &- \\
VideoAgent~\citep{fan2024videoagent} & GPT-4&  62.8&60.2 & &  -&-&-&- & &- & & - & &- & - & - & - & - \\
MoReVQA~\citep{min2024morevqa} & PaLM-2 &  -&51.7 & &  64.6&70.2&-&69.2 & &45.3 & & - & &- & - & - & - & -\\
IG-VLM~\citep{kim2024image} & GPT-4V &  -&59.8 & &  63.6&69.8&74.7&68.6 & &58.4 & & {64.2} & &- & - & - & - & -\\
VideoTree~\citep{wang2024videotree} & GPT-4 & 66.2 & 61.1 & & 70.6 & 76.5 &  83.9 & 75.6 & & - & & {66.9} & & 38.1 & 11.3 & 42.6 & 9.8 & 31.9\\ 
{LVNet~\citep{park2024too}} & {GPT-4o} & - & {61.1} & & {65.5} & {75.0} & {81.5} & {72.9} & & - & & {71.7} & & - & - & - & - & -\\
SF-LLaVA~\citep{xu2024slowfastllava} & LLaVA-NeXT-34B & - & 55.8 & & - & - &  - & 72.0 & & 55.5 & & 66.5 & & - & - & - & - & -\\ 
DrVideo~\citep{ma2024drvideo} & GPT-4 & 66.4 & 61.0 & & - & - &  - & - & & - & & - & & - & - & - & - & -\\ 
\midrule
\model{} (Ours) with LLaVA-OV & LLaVA-OV-72B & 65.0 & 63.2 & & \textbf{77.9} & \textbf{82.1} & \textbf{86.1} & \textbf{81.4} & & \textbf{64.6} & & 72.0 & & - & - & - & - & - \\ 

\model{} (Ours) with Gemini 1.5 & Gemini 1.5 Pro & 67.9 & 66.0 & & 68.3 & 72.1 & 73.4 & 71.1 & & 56.4 & & 67.9 & & \textbf{47.9} & 16.4 & 46.0 & 16.8 & \textbf{40.5}  \\ 
\model{} {(Ours) with GPT-4} & {GPT-4} & {67.5} & {66.7} & & {68.8} & {76.1} & {72.5} & {73.2} & & {59.8} & & 69.3 & & - & - & - & - & -\\ 

\model{} (Ours) with GPT-4o & GPT-4o & \textbf{69.4} & \textbf{68.1} & & 69.2 & 77.4 & 70.8 & 73.8 & & 61.3 & & \textbf{72.2} & & 38.8 & \textbf{28.7} & 64.8 & \textbf{26.1} & 37.3\\ 
\bottomrule
\label{tab:main_results}
\end{tabular}}
\vspace{-10pt}
\end{table*}

\subsection{Baselines}
Our first baselines are the flagship VLM (GPT-4o family, Gemini 1.5 family, and open-sourced LLaVA-OneVision family.). We use $1$ fps sampling to extract frames, a standard way to perform video question answering. We then use the questions, all possible answers, and the frames as prompts to the VLM to reproduce the results. The details can be found in Section \ref{sec:implementation_details}. Below we introduce the baseline that focuses on frame selection for VLM inference on video tasks. \textbf{VideoAgent} \citep{wang2024videoagent}  iteratively retrieves a new video frame from an initial uniformly sampled frame set based on high relevance until a module deems there is sufficient information to answer the query. \textbf{VideoAgent} \citep{fan2024videoagent} is another agent-based model that represents videos as a temporal memory that stores descriptions of short video segments plus an object memory storing occurrences of objects and persons, and queries both temporal and object memory to retrieve video segments. \textbf{MoReVQA} \citep{min2024morevqa} uses modularity and multistage planning, by parsing the question and using multiple LLMs for grounding, reasoning, and prediction LLM. \textbf{IG-VLM} \citep{kim2024image} transforms videos into a series of images in a six-image grid layout and use the combined grid as the VLM input. \textbf{VideoTree} \citep{wang2024videotree} recursively clusters visual embeddings of video frames,  captions them, and scores the frame relevance to build a tree structure, then traverses the tree and concatenating captions as the LLM input. We implemented VideoTree on the CLEVRER dataset based on the codebase. \textbf{LVNet} \citep{park2024too} is a framework with several hierarchical frame selection submodules that progressively filter frames to find more performance-oriented ones. \textbf{SlowFast-LLaVA} \citep{xu2024slowfastllava} considers two pathways for VLM input: a slow pathway that extracts visual features at a low frame rate for spatial details, and a fast pathway at a high frame rate to preserve temporal dynamics. \textbf{DrVideo} \citep{ma2024drvideo} converts videos into an initial document consisting of frame captions, retrieves the top K relevant captions by semantic similarity, and then employs an agent with planning to iteratively augment the document with captions from new frames until there is sufficient information to answer the question. 

\subsection{Implementation Details}
\label{sec:implementation_details}

In the following, we list descriptions of how we setup and run different VLM inferences on different datasets. As \model{} is agnostic to VLM, we evaluated \model{} on three VLMs: Gemini 1.5 family, GPT-4o family, and LLaVA-OneVision family. For all datasets except CLEVRER, we use $n=32$ frames for VLM inference. For CLEVRER, we use $n=5$ as each video is only $5$-seconds long and we adopt the standard $1$ fps. For IntentQA, we use $n=12$ for a fair comparison with baseline LVNet. We resize all videos to 720 x 480 resolution as CogVideoX only supports this format. More concretely, the initial 32 frames are encoded with the pre-trained 3D Causal VAE. Noise is added to the initial latents based on strength and the current timestep, and scaled according to the DPM scheduler’s noise sigma. The prompt texts (containing all questions and possible answers),  the noisy latents, the time step embeddings, and the rotary positional embeddings are the conditioning signals.  At denoising, the pre-trained transformer layers from CogVideoX will predict the noise and the latents are updated iteratively based on inference steps. We use 50 steps of denoising and a scale factor of 1.15258426. No masking is used. CogVideoX outputs $8N + 1$ frames, for ease of exposition we show $8N$ in ablation study tables. Then, the updated latents from the diffusion model will be leveraged by RIFE to interpolate: the latents are chunked and for each chunk, a latent frame is interpolated with the guidance of SSIM, whose threshold we can adjust to reach the number of frames. After interpolation, we have all the latent frames. 

\begin{table}[htbp]
    \centering
    \caption{
    Default number of frames for baseline methods and VAP models in Table \ref{tab:main_results}, Figure \ref{fig:performance}, and Figure \ref{fig:performance_frame_selection}. Baseline model frame varies to follow the best results from prior work \citep{li2024llava} or the standard 1 fps sampling \citep{reid2024gemini, gpt-4o}.
    }
    \label{tab:num_frames_default}
    \begin{adjustbox}{max width=0.48\textwidth}
    \begin{tabular}{ccccccc}
        \toprule
        Models & EgoSchema & NExT-QA & ANet-QA & IntentQA & CLEVRER \\
        \midrule
        GPT models (1fps) & 180 & 44 & 90 & 44 & 5 \\ 
        Gemini models (1fps)  & 180 & 44 & 90 & 44 & 5 \\
        LLaVA-OV models \citep{li2024llava} & 32 & 32 & 32 & 32 & - \\
        VAP models (Ours) & 32 & 32 & 32 & 32 & 5 \\
        \bottomrule
    \end{tabular}
    \end{adjustbox}
    \vspace{-12pt}
\end{table}

For inference, we use three models, Gemini \citep{reid2024gemini}, GPT-4o \citep{gpt-4o} and LLaVA-OneVision \citep{li2024llava}. For Gemini, Google API takes video as direct input, with default top $p$ and temperature values. Mandatory security filtering may filter out some answers and results slightly differ from the official report \citep{reid2024gemini}. For GPT-4o \citep{gpt-4o}, we use OpenCV to extract frames from videos (1 fps for standard uniform sampling baselines), and use the default top $p$ and temperature values. Non-adjustable safety settings may filter out some answers and results of GPT-4o may be slightly different from the official blog \citep{gpt-4o}. LLaVA-OneVision (LLaVA-OV) \citep{li2024llava} achieved state-of-the-art results on single-image, multi-image and video tasks, with strong transfer learning performances. The results of LLaVA-OneVision are evaluated from the LMMS framework \citep{lmms_eval2024}, following the official guideline. We do not include LLaVA-OV results on CLEVRER because the LMMS dataset containing CLEVRER data is MVBench \citep{li2024mvbench}, but only with a subset of CLEVRER, making it hard to compare with full-set CLEVRER results from other models. 

\subsection{Results}
\label{sec:results_full_names}

\paragraph{Main results.} We show results of \model{} in Table \ref{tab:main_results}. On EgoSchema, \model{} achieves $69.4\%$ accuracy on the subset and $68.1$ on the entire test set, outperforming reproduced flagship VLM baselines, as well as previous work on frame selection for VLMs. On NExT-QA, \model{} achieves higher accuracies than all baselines on all questions types: temporal (Temp.), causal (Cau.) and descriptive, with a state-of-the-art average accuracy of $81.4\%$. On ActivityNet-QA (ANet-QA), \model{} achieves state-of-the-art VideoTree results of $64.6\%$. On CLEVRER, \model{} achieves better accuracies on descriptive (Des.), explanatory (Exp.), counterfactual (Cou.), and state-of-the-art zero-shot accuracy with $40.5\%$. This shows that \model{} outperforms previous baselines by selecting effective keyframes for QA tasks than standard VLMs or recent SOTA frame selection methods. 

\paragraph{Reasoning tasks.} In particular, \model{} performs better on tasks requiring strong reasoning, i.e., temporal and causal questions in NExT-QA and explanatory and counterfactual questions CLEVRER. These questions focus on reasoning over the whole video dynamics and causes and consequences of person or object interactions. Answering these questions requires a model to select contextually important keyframes of pivotal actions leading to the outcomes. \model{} outperforms previous baselines, including standard VLMs and frame selection methods, and demonstrates its ability to select such consequential frames. 

\paragraph{{Very long video benchmark results.}}
We also show results on very long videos. We choose VideoMME \citep{fu2024video} and MLVU \citep{zhou2024mlvu}, which contain videos longer than three minutes with diverse video types, multiple video durations, and breadth in modalities. We choose the long split of VideoMME, which contains the longest 300 videos in the dataset, with the average length being 44 minutes, ranging from 30-60 minutes. We use GPT-4o mini and GPT-4o based on the LLMS-eval codebase. From Table \ref{tab:videomme}, VAP demonstrates better results than baseline GPT-4o, suggesting its effectiveness in choosing informative frames over very long video dynamics. We believe as the video generation model advances, our method can be empowered by future VLMs with stronger reasoning capacity, especially those demonstrating better spatio-temporal reasoning over very long videos.

\begin{table}
\begin{adjustbox}{max width=0.45\textwidth}

\begin{tabular}{@{}llccccc@{}}
\toprule
\multirow{2}{*}{\textbf{Model}} & \multirow{2}{*}{\textbf{VLM}} & \multirow{2}{*}{\textbf{Num. Frames}} & \multicolumn{3}{c}{\textbf{VideoMME}}            & \multirow{2}{*}{\textbf{MLVU}} \\ \cmidrule(lr){4-6}
                                &                               &                                       & \textbf{Short} & \textbf{Medium} & \textbf{Long} &                                \\ \midrule
GPT-4o-mini                     & GPT-4o-mini                   & 32                                    & 47.2           & 42.1            & 38.4          & 52.1                           \\
GPT-4o                          & GPT-4o                        & 32                                    & 70.7           & 60.1            & 54.1          & 57.8                           \\
VAP                             & GPT-4o-mini                   & 32                                    & 47.6           & 44.1            & 40.6          & 54.9                           \\
VAP                             & GPT-4o                        & 32                                    & \textbf{72.0}  & \textbf{61.8}   & \textbf{55.7} & \textbf{60.3}                  \\ \bottomrule
\end{tabular}
\end{adjustbox}
\caption{VideoMME and MLVU accuracies ($\%$). VAP demonstrates better results than baseline GPT, suggesting its effectiveness in choosing informative frames over very long videos.}
\label{tab:videomme}
\vspace{-8pt}
\end{table}





\begin{figure*}
\centering
\includegraphics[scale=0.213]{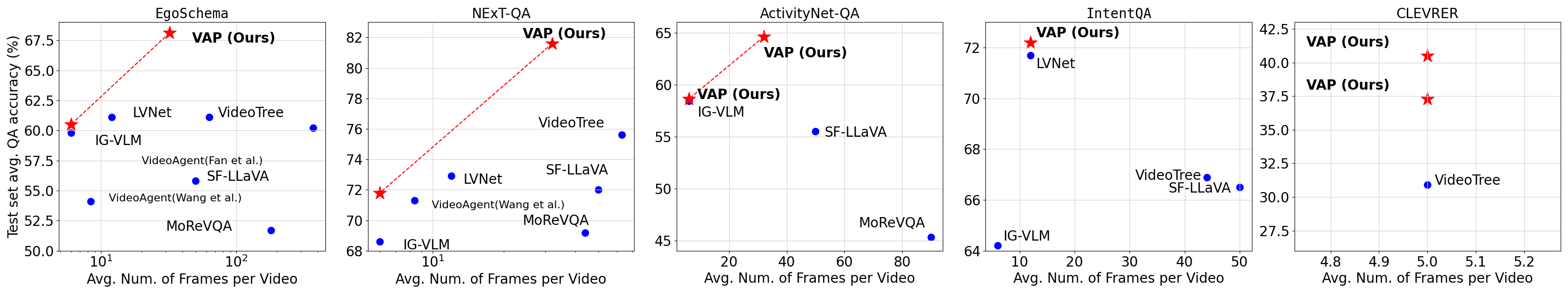}
\caption{\model{} also outperforms frame selection baselines with multiple number of frames when compared to other baselines. }
\label{fig:performance_frame_selection}
\vspace{-4pt}
\end{figure*}

\paragraph{Frame efficiency.} We showed the performance with respect to the number of frames used for the standard flagship VLMs in Figure \ref{fig:performance} in introduction. \model{} achieves better results using $n=32$ frames in EgoSchema, NExT-QA, ActivityNet-QA than standard flagship VLMs with $180$, $44$ and $90$ frames, providing $5.6 \times$, $1.5 \times$, and $2.8 \times$ frame efficiency improvements, respectively. On CLEVRER, \model{} achieves better results using the same $n=5$ frames with baselines. These results show that \model{} greatly improves the efficiency of long-form video QA for the flagship VLM inference. Next, in Figure \ref{fig:performance_frame_selection}, we also show the performance compared to other frame selection baseline methods. Because some baseline methods select fewer than $n=32$ frames, we also provide results with $n=6$.  \model{} outperforms recent frame selection baselines when compared to those with the same or a smaller number of selected frames, suggesting that \model{} is more proficient in selecting the most relevant frames compared to other baselines.

\begin{table}[htbp]
    \centering
    \caption{
    Efficiency comparison on EgoSchema between standard GPT-4o mini and VAP based on iso-compute (same latency, compare accuracies) and iso-accuracy (same accuracy, compare latencies) comparisons. VAP achieves better performance with about the same latency (VAP being slightly faster), and lower latency under the same performance. 
    }
    \label{tab:efficiency_same_performance}
    \begin{adjustbox}{max width=0.45\textwidth}
    \begin{tabular}{lccc}
        \toprule
        \textbf{Model}   & \textbf{Frames} & \textbf{Latency (seconds)} & \textbf{Accuracy} \\ \midrule
        \rowcolor{gray!20} \multicolumn{4}{c}{{\textit{Iso-compute comparison: same latency, compare accuracy}}}   \\
        GPT-4o mini & 32 & 127.2 & 56.1 \\
        VAP w/ GPT-4o mini (ours) & 16 & 125.5 & \textbf{58.6}\\
        \rowcolor{gray!20} \multicolumn{4}{c}{{\textit{Iso-accuracy comparison: same accuracy, compare latency}}}   \\
        GPT-4o mini & 96 & 232.5 & 63.3 \\
        VAP w/ GPT-4o mini (ours) & 32 & \textbf{190.4} & 63.3 \\
        \bottomrule
    \end{tabular}
    \end{adjustbox}
    \vspace{-10pt}
\end{table}

\paragraph{Computational efficiency.} We report the latency of VAP compared to standard GPT-4o mini on EgoSchema. We show two types of comparisons: iso-compute comparison, where VAP and GPT baseline share the same latency, and iso-accuracy comparison, where VAP and GPT baseline share the same accuracy.
For latency, we measure the \textit{single-request latency} by processing one video and measuring the wall-clock time from start to finish. The inference is run five times, and we report the mean. From Table \ref{tab:efficiency_same_performance}, VAP achieves better performance with about the same latency (VAP being slightly faster), and lower latency under the same performance.

\subsection{Quantitative Analyzes}
\begin{wraptable}{r}{0.25\textwidth}
    \centering
    \caption{
    EgoSchema and ActivityNet results by different numbers of selected frames (with GPT-4o mini). 
    }
    \label{tab:num_frames_ablation}
    \begin{adjustbox}{max width=0.25\textwidth}
    \begin{tabular}{cccc}
        \toprule
        \multirow{2}{*}\textbf{{Frames}} & \multicolumn{2}{c}\textbf{{Accuracy ($\%$)}}  \\
        \cmidrule(lr){2-3} 
        & \textbf{EgoSchema} & \textbf{ANet-QA} \\
        \midrule
        8 & $54.9$ & $51.5$ \\ 
         16  & $58.6$ & $53.3$\\
         32 & $63.3$ & $57.8$\\
         48 & $63.5$ & $57.7$\\
        \bottomrule
    \end{tabular}
    \end{adjustbox}
\end{wraptable}

\paragraph{Varying number of selected frames.} We are curious whether increasing the number of selected frames will increase performance. We conduct experiments on EgoSchema and ActivityNet-QA, with a number of frames ranging from $n=8$ to $n=48$, the maximum number of frames we can select due to generation capacity of the video generation model CogVideoX. Due to cost, we select the GPT-4o mini as the VLM for this comparison. We report the results in Table \ref{tab:num_frames_ablation}. The performance on \model{} with GPT-4o mini consistently improves when number of frames $n$ increases from $8$ to $32$, and plateaus when frame is greater than $n=32$, therefore we report $n=32$ results for other sections of the paper. Future work may explore using more frames for more capable models such as GPT-4o or Gemini 1.5. The results suggest that $n=32$ is optimal for \model{} and additional visual frames may be redundant or unnecessary for the questions.

\paragraph{Importance of most surprising frame selection.} We investigate whether selecting surprising frames is always beneficial even for tasks requiring vanilla video dynamics. We compare selecting the most (VAP) versus least surprising frames, and show in Appendix Table 9 that the former consistently outperforms, indicating that VAP’s strategy is effective and generalizes to diverse scenarios.

\paragraph{Reasoning abilities compared to frame selection baselines.} \label{sec:better_reasoning_over_prior_art} We are also interested in the reasoning capabilities on CLEVRER, which comprises simple rendered objects but includes challenging Explanatory (Exp.), Predictive (Pre.) and Counterfactual (Cou.) tasks. These can be challenging because of the complex multi-scene object interactions. From Table \ref{tab:main_results}, \model{} have $154.0\%$, $71.43\%$, and $8\%$ relative performance gains over VideoTree on explanatory, counterfactual, and predictive tasks, respectively, demonstrating the benefit of directly using visual frames.

\begin{wraptable}{r}{0.25\textwidth}
    \centering
    \caption{
    EgoSchema and ActivityNet results by different numbers of initial frames (with GPT-4o mini). 
    }
    \label{tab:num_initial_frames_ablation}
    \begin{adjustbox}{max width=0.25\textwidth}
    \begin{tabular}{cccc}
        \toprule
        \multirow{2}{*}{Frames} & \multicolumn{2}{c}{Accuracy ($\%$)}  \\
        \cmidrule(lr){2-3} 
        & \textbf{EgoSchema} & \textbf{ANet-QA} \\
        \midrule
        8 & $38.8$ & $34.6$ \\ 
         16  & $51.7$ & $53.3$\\
         32 & $63.3$ & $57.8$\\
         64 & $63.8$ & $58.4$\\
         80 & $63.2$ & $57.2$\\
        \bottomrule
    \end{tabular}
    \end{adjustbox}
\end{wraptable}

\paragraph{Robustness to varying initial frames.} Our next comparison focuses on varying the number of initial frames fed to the generation model, and the model's robustness to different sets of possible initial frames. This is different from the last comparison, which focuses on the number of final selected frames for VLM inference.  We conducted experiments on EgoSchema and ActivityNet-QA, with the number of frames ranging from $n=8$ to $n=80$. We also select GPT-4o mini as the VLM for this comparison. We report the results in Table \ref{tab:num_initial_frames_ablation}. The performance on \model{} with GPT-4o mini consistently improves when the number of frames $n$ increases from $8$ to $32$, stays very close from $n=32$ to $n=64$, and drops slightly at $n=80$. The first increase may suggest that more initial frames may provide the full context and dynamics of the video. However, as more frames are available, redundant frames can confuse \model{} and make it les likely to select pivotal frames that change scenes. In this paper, we use $n=32$ for all tasks except CLEVRER for ease of implementation and computation. To assess VAP’s robustness against different sets of initial frames by shifting them left and right. As shown in Appendix Sec. A.6 and Table 7, VAP’s performance is consistent, indicating minimal impact from initial frames when sampled uniformly.

\begin{table}
\centering
\begin{adjustbox}{max width=0.3\textwidth}
\begin{tabular}{@{}llcccc@{}}

\toprule

\multirow{2}{*}{\textbf{Model}} & \multirow{2}{*}{\textbf{\begin{tabular}[c]{@{}l@{}}Num.\\ Frames\end{tabular}}} & \multicolumn{4}{c}{\textbf{NExT-QA}}                          \\ \cmidrule(l){3-6} 
                                &                                                                                 & \textbf{Tem.} & \textbf{Cau.} & \textbf{Des.} & \textbf{Avg.} \\ \midrule
GPT-4o                 & 32                                                                              & \textbf{69.6} & 68.9          & 70.6          & 69.4          \\
GPT-4o                 & 44                                                                              & 67.6          & \textbf{71.9}          & \textbf{75.8} & \textbf{71.2}          \\
GPT-4o                 & 48                                                                              & 68.3          & 70.1          & 67.8          & 69.2          \\
GPT-4o                 & 64                                                                              & 67.8          & 70.0          & 68.5          & 69.1          \\
\bottomrule
\end{tabular}
\end{adjustbox}
\caption{GPT-4o baseline on NExT-QA performance plateaus at 44 frames. Selecting key frames--as done by VAP--is crucial not only for efficiency but also for achieving optimal performance.}
\vspace{-10pt}
\label{tab:increasing_num_frames_plateaus}
\end{table}
\paragraph{Increasing number of frames plateaus on baselines.} As shown in Table \ref{tab:increasing_num_frames_plateaus}, the performance of GPT-4o on NExT-QA when increasing the number of uniformly sampled frames plateaus at 44 frames, and further increases decreases performance. These results further indicate that uniform sampling frames are a suboptimal frame selection strategy. 

\paragraph{Generalization with a different generation model.} We also show results using the CogVideoX-2B model, a smaller version of the 5B model, to test the generalizability of our method. From Table \ref{tab:2b_model_result}, VAP shows performance improvement with the CogVideoX-2B model, suggesting the generalization capability of VAP.

\begin{table}
\begin{adjustbox}{max width=0.45\textwidth}

\begin{tabular}{ccccc}
\toprule
\textbf{Model} & \textbf{Model} & \textbf{EgoSchema Acc.}  & \textbf{MLVU Acc.} \\
\midrule
GPT-4o-mini                     & GPT-4o-mini                   & 56.1    & 48.6                       \\
VAP w/ CogVideoX-2B                         & GPT-4o-mini                        & 59.3    & 51.2       \\
\midrule 
LLaVA-OV-72B                             & LLaVA-OV-72B                   & 60.3    & 63.8       \\
VAP w/ CogVideoX-2B                            & LLaVA-OV-72B               & \textbf{61.4}  & \textbf{65.7} \\ \bottomrule
\end{tabular}
\end{adjustbox}
\caption{Accuracies on EgoSchema and MLVU with the smaller CogVideoX-2B. All models here take 32 frames. VAP demonstrates consistent improvements, suggesting the method generalizes over different generation models.}
\label{tab:2b_model_result}
\vspace{-6pt}
\end{table}

\subsection{Qualitative analyzes} \label{sec:qualitative}
We provide qualitative analyses on the frames selected by \model{} on EgoSchema and CLEVRER, as illustrated in Appendix Sec. A.7 and Appendix Figure 1. On EgoSchema, \model{} effectively identifies both early and late frames containing pivotal actions (e.g., taking out a phone or the final scenes that are crucial for answering the question. On CLEVRER, it similarly pinpoints collisions and trajectories necessary for explanatory and counterfactual reasoning, demonstrating its effectiveness in capturing key transitional moments across visually uninformative frames.

\section{Conclusions}
We presented Video Active Perception (\model{}), a new training-free method that significantly improves both the effectiveness and efficiency of frames for long-form video QA. \model{} achieves state-of-the-art zero-shot performance on long-form video datasets such as EgoSchema, NExT-QA, ActivityNet-QA, IntentQA, and CLEVRER. It also demonstrates better reasoning capabilities than previous methods. These results show the empirical effectiveness of \model{} and the potential of integrating prior world knowledge from diffusion models for better data acquisition and selection during VLM inference. 
\paragraph{Acknowledgement} The research is supported by National Institutes of Health awards R01MH125740, R01MH132225, R21MH130767, ONR N000142312368, and ONR MURI N00014-25-1-2116.

{
    \small
    \bibliographystyle{ieeenat_fullname}
    \bibliography{main}
}

\clearpage
\setcounter{page}{1}
\maketitlesupplementary

\section{Appendix}

\subsection{\model{} implementation details} \label{sec:app_prompt}
We provide the main prompt for the video generation model in \ref{tab:vpa_gpt4}. We provide a detailed prompt with examples to leverage the video generation model's capacity as much as possible.

In terms of the video generation model, including the video diffusion model, 3D VAE, and interpolation model, we refer the details of the model checkpoints, training and inference processes to CogVideoX.

\begin{table*}[htbp]\centering
\begin{minipage}{\textwidth}
\centering
\begin{tcolorbox} 
    \centering
   
      \scriptsize
    \begin{tabular}{p{\textwidth}} 
System: You are an advanced video generation model designed to predict plausible future video dynamics based on limited input. Your primary goal is to use your extensive prior knowledge of the world to generate latent representations of how the video is expected to unfold, given:

A few initial frames from the video;

A question about the video;

Possible answers to the question;

These generated dynamics will assist in identifying key frames in the actual video that are most informative for answering the question. \\ \\

User: Your Task: \\

1) Analyze the Initial Frames:

1a) Examine the provided initial frames to understand the setting, context, characters, objects, and any ongoing actions or events.

1b)Extract visual cues that indicate the environment (e.g., indoor, outdoor, time of day) and participants (e.g., people, animals, objects).
\\
2) Incorporate the Question and Possible Answers:

2a) Read the question carefully to determine what information is being sought.

2b) Consider each possible answer to understand different potential outcomes or scenarios.

2c) Use this information to guide your expectations of how the video might progress.
\\
3) Generate Expected Video Dynamics:

3a) Using your prior knowledge and the initial frames, predict plausible sequences of events that align with the context and are relevant to the question.

3b) Focus on generating dynamics that would lead to scenarios described in the possible answers.

3c) Create latent representations that capture these expected continuations, including scenes, events, actions, and transitions.

\\

Input Information:

1) Question about the video: $\{$Question$\}$

2) Possible Answers: $\{$Answers$\}$

3) Initial Frames: $\{$Initial Frames$\}$

\\

Instructions:

1) Leverage Prior Knowledge:

1a) Utilize your understanding of real-world behaviors, cause-and-effect relationships, and typical sequences of events.
1b) Incorporate common sense and logical reasoning to predict what is likely to happen next.

2) Focus on Relevance:

2a) Ensure that the generated dynamics are directly relevant to the question and possible answers.
2b) Highlight events or actions that would help distinguish between the different answers.

3) Maintain Consistency:
3a) Keep the generated content consistent with the visual information in the initial frames (e.g., same characters, objects, setting).
3b) Avoid introducing improbable elements that contradict the initial context.

\\

Example:

Initial Frames: Show a person standing at a crosswalk, waiting for the light to change.
Question: "What does the person do after the light turns green?"
Possible Answers:
"They cross the street."
"They turn around and walk away."
"They start jogging along the sidewalk."
Your Generated Dynamics Should:

Predict the likely actions following the initial frames, considering each possible answer.
Generate latent representations where:
The person crosses the street when the light turns green.
The person changes their mind and walks away from the crosswalk.
The person begins jogging along the sidewalk instead of crossing.

\\

Output Format:

Provide latent representations (in your internal format) that correspond to the expected video dynamics.
Ensure that these latents encapsulate the visual and temporal progression of events relevant to the question and answers.

\\
Additional Notes:

Attention to Detail: Capture subtle cues from the initial frames that might influence the outcome (e.g., the person's expression, items they are carrying, environmental conditions).
Diversity in Scenarios: While maintaining plausibility, consider multiple potential developments that are consistent with the possible answers.
Purpose of Generation: Remember that the goal is to identify discrepancies between expected and actual video content to select informative frames for further analysis.

    \end{tabular}
\end{tcolorbox}
\caption{GPT-4o prompt for \model{}}.
\label{tab:vpa_gpt4}
\end{minipage}
\end{table*}

\subsection{Experimental Details (ActivityNet-QA)}
The ActivityNet-QA test set contains 8000 QA with open-ended answers. For reproducing baselines, while GPT models can hold temporal context, they do not support videos directly. Hence, frames were sampled at 1 fps and provided to the GPT model. The videos were provided directly to the Gemini models. The format for the prompt is provided in \ref{tab:anet_gemini}. Following standard evaluation of ActivityNet-QA, we use GPT-3.5 to evaluate the open-ended answers. 

\begin{table*}[htbp]\centering
\begin{minipage}{0.95\textwidth}
\centering
\begin{tcolorbox} 
    \centering
   
      \small
    \begin{tabular}{p{0.95\textwidth}} 
   
   
Answer the following question about the video using only a word or two. Never say "unknown", "N/A" or "unsure", instead provide your most likely guess. Note that "where" questions refer to locations and not relative positions. Answer binary questions with yes or no. \\
Question: \{Question\} Answer:

    \end{tabular}
\end{tcolorbox}
\caption{Gemini and GPT-4 prompt for ActivityNet}
\label{tab:anet_gemini}
\end{minipage}
\end{table*}

\subsection{Experimental Details (Next-QA)}

\begin{table*}[htbp]\centering
\begin{minipage}{0.95\textwidth}
\centering
\begin{tcolorbox} 
    \centering
   
      \small
    \begin{tabular}{p{0.95\textwidth}} 
   
   
You are provided with a video followed by a question and choices. Answer the questions providing only the number of the correct choice. \\
\{Video\} \{Question\} 0. \{Choice 0\} 1. \{Choice 1\} 2. \{Choice 2\} 3. \{Choice 3\} 4. \{Choice 4\}
    \end{tabular}
\end{tcolorbox}
\vspace{-5pt}

\caption{Gemini prompt for Next-QA}
\label{tab:nextqa_gemini}
\end{minipage}
\vspace{-5pt}

\end{table*}

\begin{table*}[htbp]\centering
\begin{minipage}{0.95\textwidth}
\centering
\begin{tcolorbox} 
    \centering
   
      \small
    \begin{tabular}{p{0.95\textwidth}} 
   
   
These are frames from a video that I want to upload. Answer the questions providing only the number of the correct choice. \\
\{Video\} \{Question\} 0. \{Choice 0\} 1. \{Choice 1\} 2. \{Choice 2\} 3. \{Choice 3\} 4. \{Choice 4\}
    \end{tabular}
\end{tcolorbox}
\caption{GPT-4o prompt for Next-QA}
\label{tab:nextqa_gpt4}
\end{minipage}
\end{table*}

We test the reasoning capabilities of the current state-of-the-art vision language models in a zero-shot setting. The test set contains about 8500 multi-choice QA with five canditate options. For reproducing baselines, while GPT models can hold temporal context, they do not support videos directly. Hence, frames were sampled at 1 fps and provided to the GPT model. The videos were provided directly to the Gemini models. The format for the prompt is provided in \ref{tab:nextqa_gemini} for Gemini models and in \ref{tab:nextqa_gpt4} for GPT models. The prompts blocked or answered in an incorrect format (not outputting the option) by these models were dropped. The drop rate for each model is provided in \ref{tab:drop_rate}.

\begin{table}[]
\vspace{-5pt}
    \centering
    \begin{tabular}{lc}
    \hline
         \textbf{LLM} & \textbf{Drop Rate} ($\%$) \\ \hline
         GPT-4o mini& 0.7\\
         GPT-4o & 2.0\\
         Gemini 1.5 Flash & 4.2\\
        Gemini 1.5 Pro & 4.7\\ \hline
    \end{tabular}
    \vspace{-5pt}
    \caption{The percentage of points dropped for each model during evaluation due to the model blocking prompts or not answering the multiple choice.}
    \vspace{-5pt}
    \label{tab:drop_rate}
\end{table}

\begin{table}[]
\vspace{-5pt}

\centering
\begin{tabular}{@{}lc@{}}
\toprule
\textbf{Model}   & \textbf{No Answer Rate} \\ \midrule
GPT-4o           & 0.016                   \\
GPT-4o-mini      & 0.017                   \\
Gemini 1.5 Pro   & 0.142               \\
Gemini 1.5 Flash & 0.032                   \\ \bottomrule
\end{tabular}
\vspace{-5pt}

\caption{Proportion of CLEVRER multiple choice questions where no options were selected.}
\label{tab:clevrer_no_answer_rate}
\vspace{-5pt}

\end{table}

\begin{table}[]
\vspace{-5pt}
\centering
\resizebox{0.5\textwidth}{!}{

\begin{tabular}{@{}lcccc@{}}
\toprule
\textbf{Model} & \textbf{VLM}   & \textbf{Set of Frames} & \textbf{EgoSchema} & \textbf{Anet-QA} \\ \midrule
VAP         & GPT-4o mini & Left & 63.1 & 57.6 \\
VAP         & GPT-4o mini & Middle & 63.3 & 57.8 \\
VAP         & GPT-4o mini & Right & 63.3 & 57.7 \\
VAP         & Gemini-1.5 Flash & Left & 62.9 & 56.4 \\
VAP         & Gemini-1.5 Flash & Middle & 62.8 & 56.2 \\
VAP         & Gemini-1.5 Flash & Right & 62.8 & 56.5 \\

\bottomrule
\end{tabular}
\vspace{-5pt}
}
\caption{Shuffling initial frames. The performances are consistent with different sets of initial frames, suggesting the impact of different initial frames is minimal, as long as the initial frames are uniformly sampled.}
\label{tab:initial_frames}
\vspace{-5pt}

\end{table}

\begin{table}[htbp]
\begin{tabular}{@{}lllcl@{}}
\toprule
\multirow{2}{*}{\textbf{Model}} & \multirow{2}{*}{\textbf{VLM}} & \multirow{2}{*}{\textbf{Num. Frames}} & \multirow{2}{*}{\textbf{Acc. (\%)}} & \multirow{2}{*}{\textbf{Cost (\$)}} \\
                                &                               &                                       &                                     &                                     \\ \midrule
GPT-4o                          & GPT-4o                        & \multicolumn{1}{c}{20}                & 39.2                                & \multicolumn{1}{c}{63.8}            \\
VAP                             & GPT-4o                        & \multicolumn{1}{c}{\textbf{10}}       & \textbf{42.1}                       & \multicolumn{1}{c}{\textbf{31.9}}   \\ \bottomrule
\end{tabular}
\caption{CLEVRER results comparing GPT-4o baseline and VAP. VAP outperforms GPT-4o with half the frames and half the monetary cost, demonstrating VAP's effectiveness both performance-wise and economically.}
\label{tab:clevrer_extra}
\end{table}

\begin{table}[htbp]
\centering
\begin{adjustbox}{width = 0.48\textwidth}
    
\begin{tabular}{@{}llcccc@{}}
\toprule
\multirow{2}{*}{\textbf{Selection}}                             & \multirow{2}{*}{\textbf{VLM}} & \multirow{2}{*}{\textbf{Frames Selected}} & \multicolumn{2}{c}{\textbf{EgoSchema}}                                  & \textbf{NExT-QA}                         \\ \cmidrule(l){4-6} 
                                                                &                               &                                           & \multicolumn{1}{l}{\textbf{Subset}} & \multicolumn{1}{l}{\textbf{Full}} & \multicolumn{1}{l}{\textbf{Descriptive}} \\ \midrule
Least surprising                                                & Gemini 1.5 Flash              & 32                                        & 53.6                                & 50.8                              & 64.1                                     \\
Least surprising                                                & GPT-4o mini                   & 32                                        & 56.3                                & 53.2                              & 62.2                                     \\
Random                                                & Gemini 1.5 Flash              & 32                                        & 61.1                                & 58.7                              & 63.8                                     \\
Random                                                & GPT-4o mini                   & 32                                        & 62.3                                & 58.9                              & 63.4                                     \\

Most surprising\\ (VAP) & Gemini 1.5 Flash              & 32                                        & 64.2                                & 62.8                              & \textbf{66.4}                            \\
Most surprising\\ (VAP)  & GPT-4o mini                   & 32                                        & \textbf{65.6}                       & \textbf{63.3}                     & 65.3                                     \\ \bottomrule
\end{tabular}
\end{adjustbox}
\caption{EgoSchema and NExT-QA results when selecting the least surprising, randomly sampled, and most surprising (VAP) frames. We use $32$ frames across datasets for consistency. The random baseline performances will differ from those in Table 1, which use $180$ frames for EgoSchema and $44$ for NExT-QA, following prior work. Results show that our proposed approach, selecting the most surprising frames, is advantageous.}
\label{tab:most_least_surprising}
\end{table}

\begin{table}
\begin{adjustbox}{max width=0.45\textwidth}

\begin{tabular}{ccccc}
\toprule
\textbf{Model} & \textbf{Model} & \textbf{EgoSchema Acc.}  & \textbf{MLVU Acc.} \\
\midrule
GPT-4o-mini                     & GPT-4o-mini                   & 56.1    & 48.6                       \\
VAP                         & GPT-4o-mini                        & 60.6    & 52.0       \\
\midrule 
LLaVA-OV-72B                             & LLaVA-OV-72B                   & 60.3    & 63.8       \\
VAP                            & LLaVA-OV-72B               & \textbf{62.0}  & \textbf{66.3} \\ \bottomrule
\end{tabular}
\end{adjustbox}
\caption{Accuracies on EgoSchema and MLVU with the E-LatentLPIPS metric. All models here take 32 frames. VAP demonstrates consistent improvements on the new metric. E-LatentLPIPS also outperforms the cosine similarity, suggesting the superiority of perceptual-based metrics.}
\label{tab:LatentLPIPS}
\vspace{-6pt}
\end{table}

\begin{table*}[htbp]\centering
\begin{minipage}{0.95\textwidth}
\centering
\begin{tcolorbox} 
    \centering
   
      \small
    \begin{tabular}{p{0.95\textwidth}} 
   

You will be provided frames from a video, sampled evenly across the video. You will also be given a question about the video and an enumerated list of options. Select all options that are correct. After explaining your reasoning, output your final answer in the format "Final Answer: {comma separated list of correct option numbers}". At least one option is correct, so always pick the option(s) that are most likely to be correct even if no option seems entirely correct. \\ \\
\{\{video\}\} \\ 
Question: \{\{ question \}\} \\
Options: \\
\{\% for option in options \%\} \\
(\{\{ loop.index0 \}\}) \{\{ option \}\} \\
\{\% endfor \%\} \\ 
    \end{tabular}
\end{tcolorbox}
\caption{Prompt for CLEVRER multiple choice questions.}
\label{clevrer_mcq_prompt}
\end{minipage}
\end{table*}

\begin{table*}[htbp]\centering
\begin{minipage}{0.95\textwidth}
\centering
\begin{tcolorbox} 
    \centering
   
      \small
    \begin{tabular}{p{0.95\textwidth}} 
   
   
You will be provided frames from a video, sampled evenly across the video. Answer the question about the video using only a word or number. Never say "unknown", "N/A" or "unsure", instead provide your most likely guess. Answer binary questions with yes or no. \\ \\
\{\{video\}\} \\
Question: \{\{question\}\} Answer:  
    \end{tabular}
\end{tcolorbox}
\caption{Prompt for CLEVRER binary questions.}
\label{clevrer_word_ans_prompt}
\end{minipage}
\end{table*}

\subsection{Experimental Details (CLEVRER)}

We evaluate on CLEVRER’s test set, which contains 5,000 videos and 76,340 QA pairs. For multiple choice questions, we report both per-option accuracy and per-question accuracy. Per-option accuracy measures the overall correctness of selected options across all questions, and per-question accuracy measures the overall correctness of questions which require all choices to be selected correctly. For reproducing baselines, in both Gemini 1.5 and GPT-4o we sample the videos at 1 fps. We use the prompts in \ref{clevrer_mcq_prompt} and \ref{clevrer_word_ans_prompt} for multiple choice and single word answer questions respectively. Furthermore, we report in \ref{tab:clevrer_no_answer_rate} the proportion of multiple choice questions for which Gemini 1.5 and GPT-4o do not select any of the options, classifying them all as incorrect. For Gemini 1.5 Pro, we initially observed that no options were selected for 25.4\% of the multiple choice questions, a significantly higher rate than Gemini 1.5 Flash and both GPT-4o variants. After re-evaluating Gemini 1.5 Pro on these questions, the rate of multiple choice questions with no selected options dropped to 14.2\%. 

\begin{table*}[htbp]\centering
\begin{minipage}{0.95\textwidth}
\centering
\begin{tcolorbox}[width=\textwidth] 
    \centering
   
      \small
    \begin{tabular}{p{0.9\textwidth}} 
   
   
You will be given a question about a video and five possible answer options, where C refers to the person wearing the camera. You will be provided frames from the video, sampled evenly across the video. \\
\{video\} \\
Question: \{question\} \\ 
Possible answer choices: \\
(0) \{option0\} \\
(1) \{option1\} \\
(2) \{option2\} \\
(3) \{option3\} \\
(4) \{option4\} \\
After explaining your reasoning, output the final answer in the format "Final Answer: (X)", where X is the correct digit choice. Never say "unknown" or "unsure", or "None", instead provide your most likely guess.
    \end{tabular}
\end{tcolorbox}
\caption{Prompt for the EgoSchema dataset.}

\label{egoschema_prompt}
\end{minipage}
\end{table*}

\subsection{Experimental Details (EgoSchema)}
We evaluated EgoSchema on the entire set of 5,000 question answer pairs. Each question was multiple choice, with 5 answers. We calculated the percentage of correct multiple choice answers on the entire set of 5,000 questions. Each video in EgoSchema is 3 minutes long, which is 180 seconds. For reproducing baselines, in order to sample the video frames, we processed one frame per second, and passed in the array of 180 frames. In terms of models, we evaluated EgoSchema on GPT-4, GPT-4o, and Gemini. The specific prompt format for each of these is shown in Fig. \ref{egoschema_prompt}.

\subsection{{Robustness to initial frames}}
\label{sec:robustness}
We perform the following experiments to show that shuffling initial frames has a minimal impact on the performance. We select different initial frames on EgoSchema subset and ActivityNet-QA: we select three different sets of initial frames: [left, middle, right], where the middle is the standard VAP, and left and right sets are shifting the frame IDs left and right by one-third of a block, where a block is the number of total unsampled frames between two sampled frames. This gives two extra sets of initial frames, as uniformly distributed as the original set in VAP. We sample 32 frames from each video. We report the results in Table \ref{tab:initial_frames}. From the results, the performances are consistent with different frame sets, suggesting the impact of different initial frames is minimal, as long as the initial frames are uniformly sampled.

\subsection{E-LatentLPIPS similarity metric}
We conducted experiments on EgoSchema and MLVU with a different similarity metric: E-LatentLPIPS, a perceptual distance metric that operates directly in the latent space, utilizing an ensemble of augmentations. For ease of computation, we also use CogVideoX-2B. From Table \ref{tab:LatentLPIPS}, VAP demonstrates consistent improvements on the new metric. E-LatentLPIPS also outperforms the cosine similarity (Table 8 in main text), suggesting the superiority of perceptual-based metrics.

\subsection{Qualitative analysis} \label{sec:qualitative_app}

\begin{figure*}[h]
    \centering
    \begin{subfigure}{0.9\textwidth}
        \centering
        \includegraphics[width=0.9\linewidth]{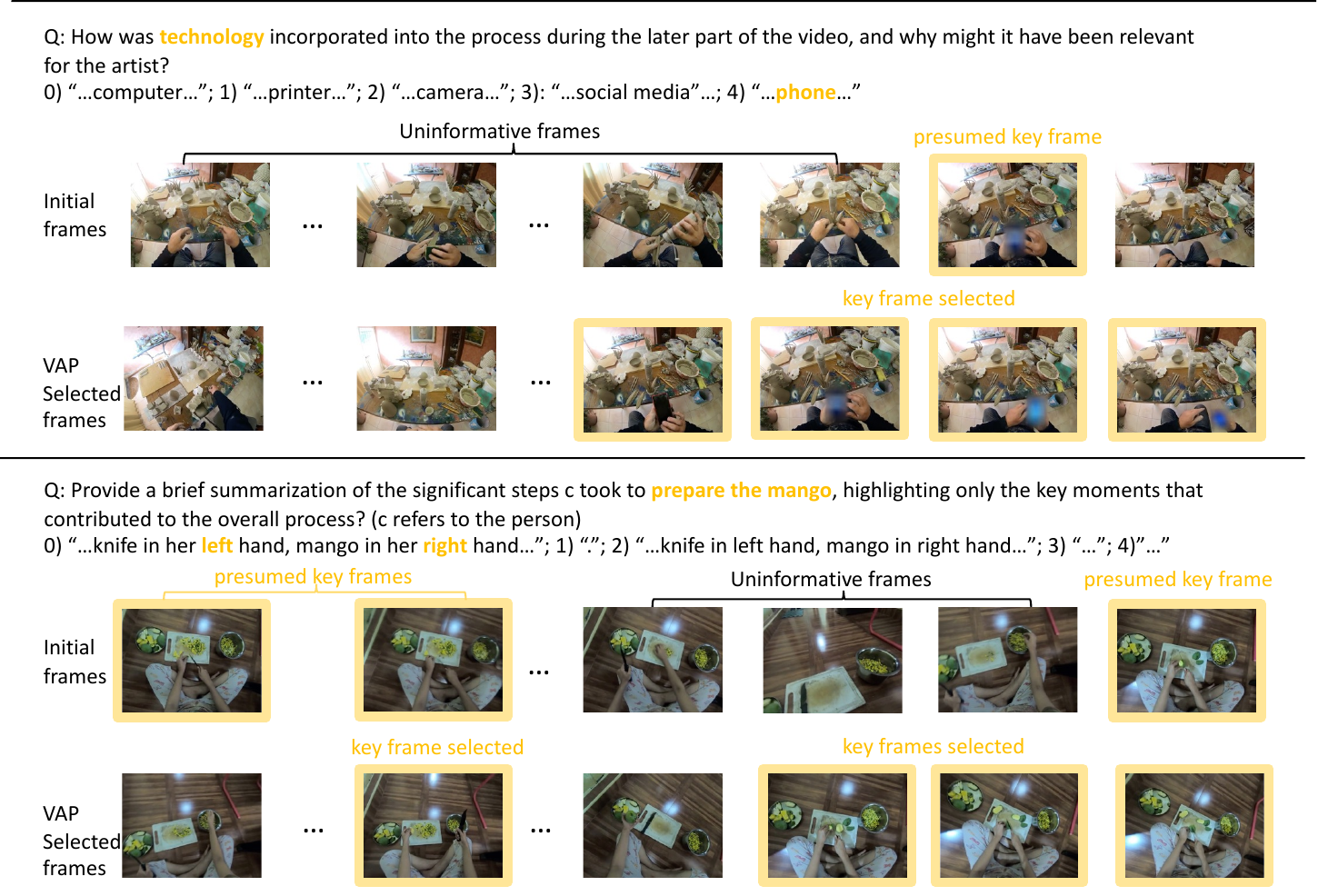} 
        \caption{Qualitative analyzes of \model{} on EgoSchema.}
        \label{fig:subfig1}
    \end{subfigure}
    \begin{subfigure}{0.9\textwidth}
        \centering
        \includegraphics[width=0.9\linewidth]{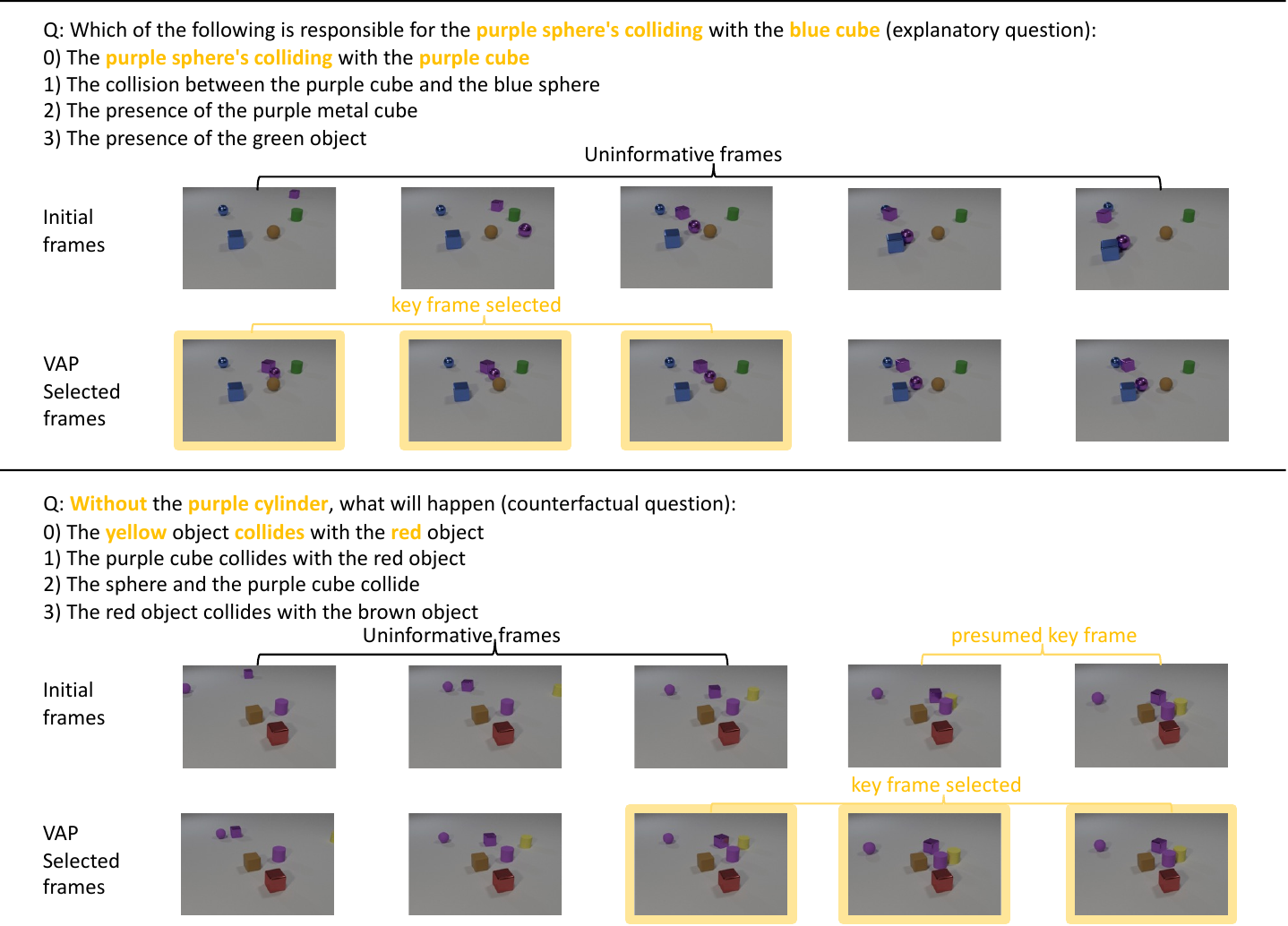} 
        \caption{Qualitative analyzes of \model{} on CLEVRER.}
        \label{fig:subfig2}
    \end{subfigure}
    \caption{Qualitative analyzes show that \model{} can select key transitioning frames relevant for question-answering, including action frames towards end of video (first example), keyframes at both the beginning and at the end of the video (second example), unseen key collision frames (third example), and frames that demonstrate the pivotal pre-collision and post-collision moments for counterfactual question (fourth example).}
    \label{fig:qualitative_examples}    

\end{figure*}

On EgoSchema (Figure \ref{fig:subfig1}), the first example requires frames towards the end of the video to answer the question, and VAP correctly selects most frames towards the end. Presumed keyframe (a person taking out a phone) from the initial frames is different from the rest of the scene but is question relevant, and \model{} are able to extract frames adjacent to this keyframe. In the second example, the question can be answered by both the beginning and the end of the video, despite the visual detail differences. \model{} is successfull at selecting the frames at both ends of the video, especially at the end of the video where the scenes are most useful for the answer. In both cases, \model{} is successful in finding frames that contain crucial tasks-relevant frames, although they differ from other uninformative frames in visual details.

On CLEVRER, the explanatory and counterfactual questions are particularly hard because they require reasoning over positions and interactions over many objects. We show two examples for each type of question (Figure \ref{fig:subfig2}). In the first CLEVRER example, a purple sphere first collides with a purple cube, and the purple sphere then collides with the blue cube. The question is to ask which object is responsible for the later collision. None of the initial frames actually shows the first collision, which is the key to the answer. \model{}, however, can successfully pick up the key scenes of the first collision, as these frames are visually different from initial frames but are question-relevant. In the second example, a yellow object outside the scene comes in and collides with the purple cylinder. The counterfactual question is to ask what happens if there is no purple cylinder (the correct answer is that the yellow object will collide with the red object). Answering this question requires frames that show the trajectories of the yellow object, which are frames containing the pivotal pre-collision and post-collision moments. \model{} successfully picked up these frames, demonstrating its effectiveness in selecting key transitioning frames.

\end{document}